\documentclass[10pt,twocolumn,letterpaper]{article}

\usepackage{cvpr}              
\usepackage{booktabs}
\usepackage{adjustbox}  

\definecolor{cvprblue}{rgb}{0.21,0.49,0.74}
\usepackage[pagebackref,breaklinks,colorlinks,allcolors=cvprblue]{hyperref}
\usepackage{pifont} 

\def\paperID{2959} 
\def\confName{CVPR}
\def\confYear{2026}

\title{Rethinking Position Embedding as a Context Controller \\ for Multi-Reference and Multi-Shot Video Generation}

\author{
Binyuan Huang$^{1}$, Yuning Lu$^{2}$\footnotemark[2], Weinan Jia$^{2}$, Hualiang Wang$^{3}$, Mu Liu$^4$, Daiqing Yang$^1$ \\
$^1$ Wuhan University \quad$^2$ University of Science and Technology of China  \\
$^3$ Hong Kong University of Science and Technology \quad $^4$ Tsinghua University   \\
{\small \texttt{by\_huang@whu.edu.cn,} \texttt{luyuningx@gmail.com}}\\
{\small Project Page: \url{https://poco-multiref-multishot.github.io/}}
}

\AtBeginDocument{
  \abovedisplayskip=4pt plus 2pt minus 2pt
  \abovedisplayshortskip=2pt plus 1pt minus 1pt
  \belowdisplayskip=4pt plus 2pt minus 2pt
  \belowdisplayshortskip=2pt plus 1pt minus 1pt
}

\begin{document}

\maketitle

\renewcommand{\thefootnote}{\fnsymbol{footnote}}
\footnotetext[2]{Corresponding authors.}
\renewcommand{\thefootnote}{\arabic{footnote}}

\begin{abstract}

Recent proprietary models such as Sora2 demonstrate promising progress in generating multi‑shot videos conditioned on multiple reference characters. However, academic research on this problem remains limited.
We study this task and identify a core challenge: when reference images exhibit highly similar appearances, the model often suffers from reference confusion, where semantically similar tokens degrade the model’s ability to retrieve the correct context.
To address this, we introduce \textbf{PoCo} (Position Embedding as a Context Controller), which incorporates position encoding as additional context control beyond semantic retrieval.
By employing side information of tokens, PoCo enables precise token‑level matching while preserving implicit semantic consistency modeling.
Building on PoCo, we develop a multi‑reference and multi‑shot video generation model capable of reliably controlling characters with extremely similar visual traits.
Extensive experiments demonstrate that PoCo improves cross‑shot consistency and reference fidelity compared with various baselines.
\end{abstract}    
\begin{figure*}[t]
\begin{center}
\includegraphics[width=0.92\linewidth]{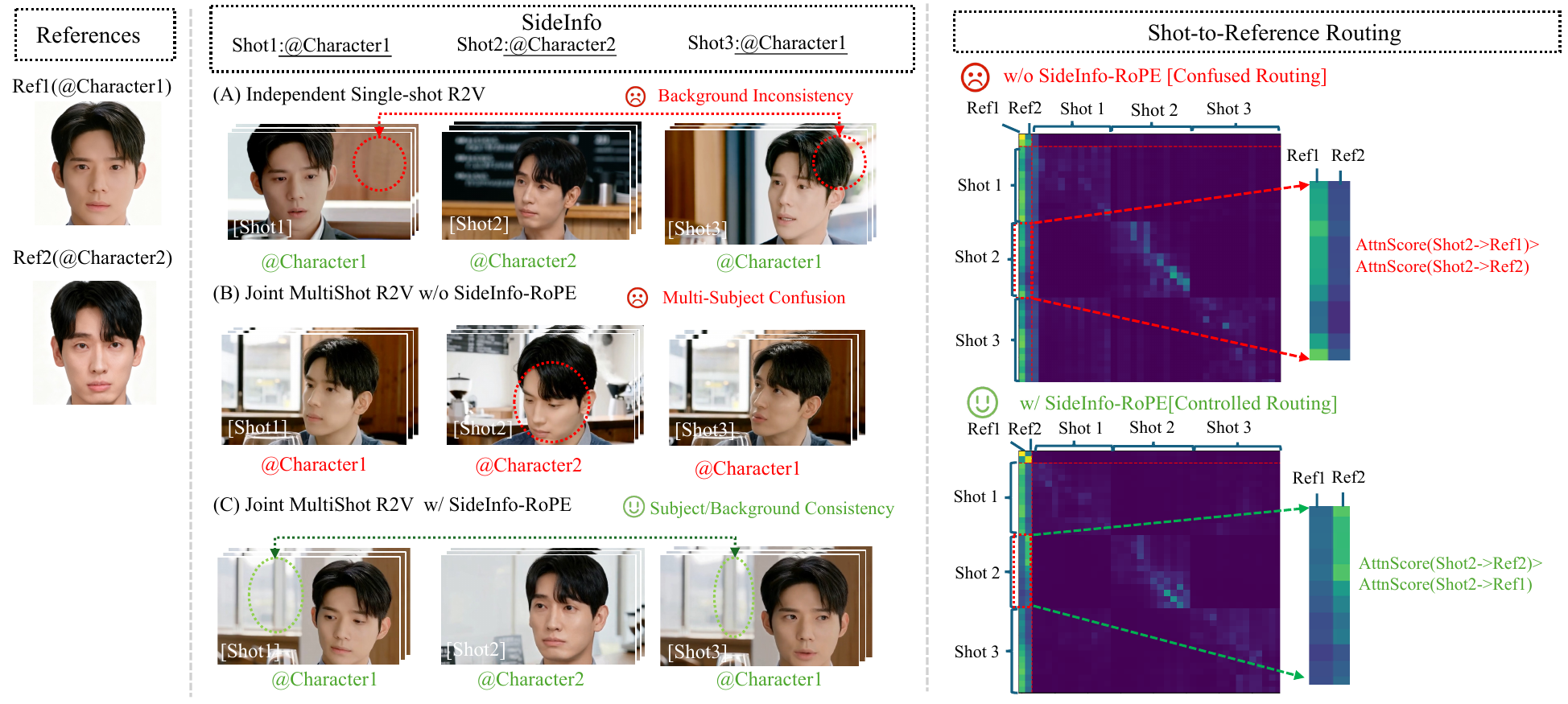}
\end{center}
\vspace{-18pt}
\caption{
\textbf{Comparison of different strategies for multi-reference, multi-shot video generation.}
(A) \textit{Independent single-shot reference-to-video generation} produces each shot separately, leading to inconsistent backgrounds and appearance details across shots.
(B) \textit{Joint multi-shot reference-to-video generation} improves global coherence, but without explicit side information, the model may associate a shot with the wrong reference, causing identity confusion.
(C) Our \textbf{PoCo} with \textbf{SideInfo-RoPE} enables accurate shot-reference association, yielding consistent identity and background across shots.
Right: spatially averaged self-attention over concatenated reference and shot tokens. \(\mathrm{AttnScore}(\mathrm{Shot}_i \rightarrow \mathrm{Ref}_j)\) denotes the mean attention from \(\mathrm{Shot}_i\) to \(\mathrm{Ref}_j\).
}
 \label{3wayscompare}
\end{figure*}

\section{Introduction}

Despite rapid advances in video generation with Diffusion Transformers (DiTs) across various tasks (\eg, text-to-video generation~\cite{yang2024cogvideox,wan2025wan, kong2024hunyuanvideo, gao2025seedance, zhang2025waver}, reference-to-video generation~\cite{liu2025phantom,jiang2025vace,li2023videogen,chen2025multi}, and multi-shot video generation~\cite{wang2025echoshot,jia2025moga,kara2025shotadapter,guo2025long}), scaling these capabilities to complex scenarios remains underexplored.
In particular, multi-reference and multi-shot video generation is crucial: it enables (1) the capture of holistic scene semantics and temporal continuity across shots, and (2) coherent generation while preserving global reference consistency, thereby enhancing both the realism and the narrative coherence of the generated videos.

While a few closed-source and technically opaque systems (\eg, OpenAI’s Sora2~\cite{openai@sora2}) demonstrate the potential of multi-shot and multi-reference video generation, robust and transparent methods in this area remain limited. Most existing approaches~\cite{liu2025phantom,jiang2025vace} rely on multiple independent reference-to-video pipelines, a restrictive paradigm that generates each shot in isolation based on its corresponding reference and textual prompt. 
This design preserves reference fidelity within individual shots, but it blocks semantic interactions between tokens across shots and undermines scene-level semantic consistency and temporal coherence, as illustrated in Fig. \ref{3wayscompare}.

A straightforward attempt is to extend existing reference-to-video methods~\cite{liew2023magicedit, liu2024video, zhang2025magicmirror} (\ie, concatenating reference and generated tokens in the context) to multi-reference, multi-shot scenes, expecting the attention mechanism within the DiT to implicitly establish correct associations between reference tokens and generated tokens. 
However, our experiments show that the model encounters \textbf{reference confusion} in such complex scenarios, as shown in Fig.~\ref{3wayscompare}. When multiple references exhibit subtle appearance differences, semantic similarity alone is often insufficient to establish reliable shot-reference associations, causing the model to retrieve information from an unintended reference and thus degrading visual fidelity.
The attention visualizations further reveal that this failure stems from incorrect shot-reference associations, with a shot attending more strongly to the wrong reference than to its intended one.

The dilemma between (1) implicitly ensuring multi‑shot scene-level consistency and (2) faithfully maintaining distinct representations of references arises from the lack of explicit, fine‑grained conditioning signals beyond native attention's semantic context retrieval, leaving the model unable to precisely control which reference information is incorporated into each shot.
\emph{Is it possible to retain the implicit scene‑level consistency enabled by semantic routing while simultaneously preventing semantics‑induced reference confusion, thereby achieving explicit consistency?}

To address this issue, we revisit the attention mechanism and decompose it into two functional components. 
The first is a learnable, semantics-driven component, in which queries retrieve relevant keys from the context. 
The second is a manually designed positional embedding that encodes spatiotemporal relations to \textit{organize} the context. 
This perspective suggests that positional embeddings can serve as an additional mechanism for context control, enabling precise reference conditioning beyond native semantic control.

To this end, we define \textbf{side information} for tokens as auxiliary attributes that complement semantic cues and facilitate more precise context retrieval. 
Concretely, in this work, we treat the extra metadata provided by the user in each shot prompt, such as reference identifiers \texttt{@character\_i}, as side information associated with the corresponding tokens.
These identifiers provide a precise and reliable signal for establishing associations among tokens belonging to the same reference.

Finally, we propose \textbf{PoCo} (\textbf{Po}sition Embedding as a \textbf{Co}ntext Controller), a solution for multi-reference and multi-shot video generation that enables precise context control.
PoCo measures the degree of match between tokens with respect to side information as a distance, which is then treated as their relative position in the side information space, allowing position embeddings to naturally organize context beyond semantic cues. 
It preserves full attention connectivity while injecting side information signals into query–key interactions, enabling the model to resolve reference ambiguity during attention.
We instantiate this idea with \textbf{SideInfo‑RoPE}, a side information‑aware extension of Rotary Position Embedding (RoPE)~\cite{su2024roformer}, which introduces an additional axis that encodes side information similarity.
PoCo provides an explicit and efficient context controller without introducing additional memory or computational overhead, while retaining the implicit semantic interaction of native attention.

In summary, our contributions are as follows:
\begin{itemize}
    \item We identify the challenge of reference confusion in multi‑shot, multi‑reference video generation, and introduce PoCo, position embedding as context controller that leverages side information of token for precise context retrieval, effectively mitigating semantic‑induced reference ambiguity.
    
    \item Building on PoCo, we develop a multi‑reference, multi‑shot video generation model capable of faithfully preserving fine‑grained visual identities, even when reference images exhibit subtle inter‑instance variations.
    \item Extensive experiments demonstrate that PoCo significantly improves cross-shot consistency and reference fidelity, outperforming both shot-segmented R2V methods.  
\end{itemize}

\section{Related Work}

\subsection{Reference-to-Video Generation}
Reference-guided video generation \cite{zhao2025controlvideo, huang2023composer,wang2023videocomposer,huang2025subjectdrive,guo2024pulid,wang2024instantid, liu2023cones} aims to synthesize videos that faithfully preserve the identity and appearance specified by reference images.
Recent work shows that incorporating visual references substantially enhances controllability and identity fidelity.
Phantom~\cite{liu2025phantom} and SkyReels~\cite{fei2025skyreels} encode reference images using the VAE encoder and inject the resulting latents into the video diffusion process alongside CLIP-based semantic conditioning.
VACE~\cite{jiang2025vace} extends this paradigm with a unified conditioning framework that leverages a context-adapter branch to flexibly incorporate reference images during generation.
HunyuanCustom~\cite{hu2025hunyuancustom} further integrates an MLLM-based text-image interaction module together with VAE-based identity conditioning, improving semantic-visual alignment.
However, these methods are primarily optimized for \textit{single-shot} video generation and lack mechanisms to ensure consistent character identity and background continuity across multiple shots.

In \textit{multi-shot} scenarios, only a limited number of methods explore reference-conditioned consistency~\citep{wang2025echoshot,qiu2025animeshooter}. EchoShot~\cite{wang2025echoshot} supports single-identity multi-shot generation, but does not explicitly study multi-reference settings.   AnimateShooter~\cite{qiu2025animeshooter} introduces an anime-focused dataset for multi-reference, multi-shot generation, with a primary focus on the animation domain.

\subsection{Video Generation with Context Control}

Video generation with DiT employs attention mechanisms for context control. 
Existing work on context control typically relies on the implicit semantics of tokens for context retrieval, often selecting salient query–key pairs to enable efficient sparse attention. 
In long video generation, some methods design token-level routing \cite{jia2025moga} or block-level retrieval~\cite{cai2025moc, wu2025vmoba} to implicitly maintain long-range consistency with long context. 
However, these approaches do not explicitly address consistency in reference-conditioned video generation.

A recent related work, “Context as Memory”~\cite{yu2025context} also employs auxiliary information for context retrieval.
The method retrieves frames from a historical memory using overlapping fields of view to identify relevant frames, enabling interactive long-video generation.
In contrast, PoCo differs in a fundamental way: it removes the need for any \textit{external} retrieval modules and performs soft control directly within the native attention. 
By leveraging position embeddings to guide control, PoCo preserves the complete context and supports efficient end-to-end generation.

\subsection{Video Generation with RoPE}
RoPE~\cite{su2024roformer} has become a central mechanism for modeling temporal structure in video diffusion models. Early approaches primarily apply 3D-RoPE to video latents to encode frame wise spatiotemporal positions under the assumption of continuous clips. Building on this idea, EchoShot~\cite{wang2025echoshot} introduces shot aware RoPE, where TcRoPE injects temporal phase shifts at shot boundaries and TaRoPE modulates cross attention to align each shot with its own caption while suppressing cross shot interference, enabling native multi shot generation. In parallel, HunyuanCustom~\cite{hu2025hunyuancustom} and Stand In~\cite{xue2025stand} extend 3D-RoPE by assigning offset temporal indices and disjoint spatial coordinates to reference image tokens, placing them at virtual negative time steps or outside the video’s spatial grid, so that reference latents remain geometrically separated yet attendable, expanding RoPE from a simple positional encoder into a flexible tool for modeling shot structure and reference based identity control.

\section{Method}
\begin{figure*}[htbp]
\begin{center}
\includegraphics[width=1.0\linewidth]{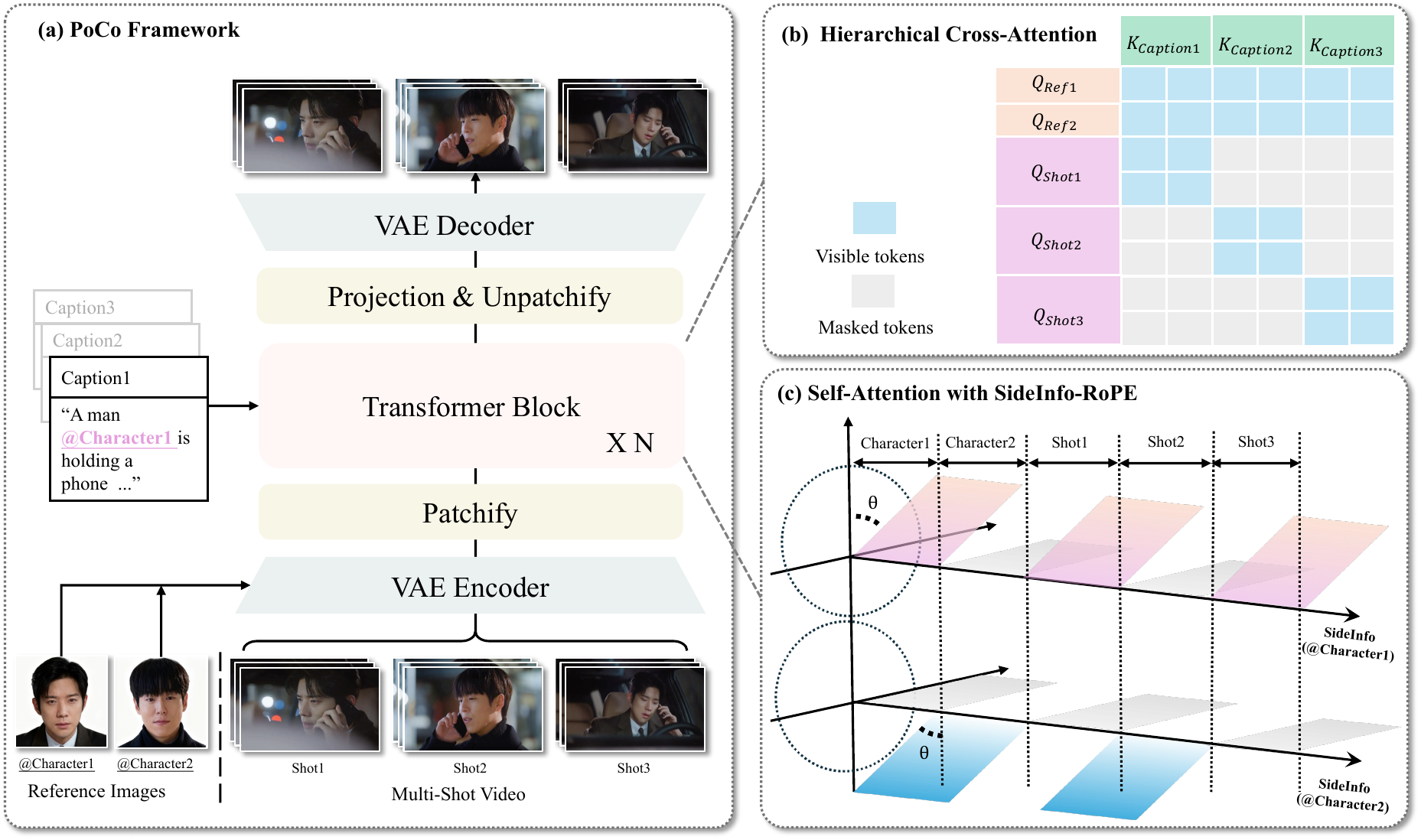}
\vspace{-18pt}
\end{center}
   \caption{
We propose a multi-reference, multi-shot video generation model conditioned on reference images and per-shot captions. 
(a) The overall architecture integrates reference images, shot captions, and latent video features through VAE and MultiShot-DiT blocks. Each block contains \textbf{Hierarchical Cross-Attention} (b) and \textbf{Self-Attention with SideInfo-RoPE} (c). 
(b) The hierarchical mask allows reference tokens to attend to all captions, while video tokens in each shot attend only to their corresponding text segment. 
(c) SideInfo-RoPE assigns reference-specific phase codes in the rotary embedding space, so that temporally aligned shots inherit the corresponding phase patterns. Colored planes denote active rotations, while gray planes denote unrotated ones.
   }
 \label{overview}
\end{figure*}

In this section, we first present the Preliminary for RoPE and then elaborate on the proposed SideInfo-RoPE and Hierarchical Cross-Attention, as illustrated in Fig.~\ref{overview}. Finally, the data pipeline is described.

\subsection{Preliminary}

\textbf{RoPE} encodes token positions by rotating query and key vectors in a complex-valued phase space, 
yielding a continuous, direction-aware representation of \emph{relative} positions within dot-product attention.
Unlike absolute sinusoidal positional encodings, 
RoPE injects relative positional information directly into the attention score, allowing it to depend jointly on content similarity and relative displacement.

We first consider the 1D sequence case.
Let $\boldsymbol{x}_m$ and $\boldsymbol{x}_n$ be tokens at positions $m$ and $n$, with corresponding query–key vectors $(\boldsymbol{q}_m, \boldsymbol{k}_n) \in \mathbb{R}^{D}$ for a single attention head of dimensionality $D$. 
The RoPE-modulated attention score between these two tokens is:
\begin{equation}
\boldsymbol{q}_m^{\top} \, \boldsymbol{R}_{\Delta_{m,n}}(\theta) \, \boldsymbol{k}_n,
\end{equation}
where $\Delta_{m,n} = m - n$ denotes the relative offset, $\theta$ is a predefined base frequency, and the rotation matrix $\boldsymbol{R}_{\Delta_{m,n}}(\theta)$ is an orthogonal block-diagonal matrix that applies a position-dependent phase to paired coordinates.
For brevity, we write $\boldsymbol{R}_{\Delta_{m,n}}$ when $\theta$ is clear from context. 
Specifically,
\begin{equation}
\boldsymbol{R}_{\Delta_{m,n}} \;=\; \bigoplus_{i=1}^{D/2} \boldsymbol{R}^{(i)}_{\Delta_{m,n}},
\end{equation}
where $\oplus$ denotes the \emph{blockwise direct sum}, and each $2\times 2$ block acts on a consecutive coordinate pair:
\begin{align}
\boldsymbol{R}^{(i)}_{\Delta_{m,n}}(\theta) & \;=\;
\begin{bmatrix}
\cos\!\big( \omega_i \cdot \Delta_{m,n} \big) & -\sin\!\big(\omega_i \cdot \Delta_{m,n}  \big) \\
\sin\!\big( \omega_i \cdot \Delta_{m,n} \big) & \phantom{-}\cos\!\big(\omega_i \cdot \Delta_{m,n}  \big)
\end{bmatrix}, \\
& \omega_i = \theta^{-\frac{2i}{D}}.
\end{align}

We extend RoPE to \textbf{3D-RoPE} for representing 3D positions in video generation.
Each token corresponds to a spatiotemporal patch with coordinates $\boldsymbol{p}=(t,h,w)$. 
For two tokens at $\boldsymbol{p}_m=(t_m,h_m,w_m)$ and $\boldsymbol{p}_n=(t_n,h_n,w_n)$, the relative displacement is:
\begin{equation}
\Delta^{\boldsymbol{p}}_{m,n} \;=\; \big(\Delta^t_{m,n},\, \Delta^h_{m,n},\, \Delta^w_{m,n}\big),
\end{equation}
with $\Delta^t_{m,n}=t_m-t_n$, etc.
We allocate disjoint subspaces of the head dimension $D$ to each axis.
Let $D_t, D_h, D_w \in \mathbb{N}$ satisfy $D_t + D_h + D_w = D$. The 3D-RoPE rotation is
{
\begin{align}
    \boldsymbol{R}_{\Delta^{\boldsymbol{p}}_{m,n}} = & \{ \bigoplus_{i=1}^{\frac{D_t}{2}} \boldsymbol{R}^{(i)}_{\Delta^t_{m,n}} \} \oplus \{ \bigoplus_{i=1+\frac{D_t}{2}}^{\frac{D_t}{2}+\frac{D_h}{2}} \boldsymbol{R}^{(i)}_{\Delta^{h}_{m,n}} \} \notag \\ \oplus &\{ \bigoplus_{i=\frac{D_t}{2}+\frac{D_h}{2}}^{\frac{D}{2}} \boldsymbol{R}^{(i)}_{\Delta^{w}_{m,n}} \},
\end{align}
}
where each component is constructed as above within its allocated subspace.

\subsection{SideInfo-RoPE}
\label{subsec:sideinfo_rope}

We observe that multi-reference and multi-shot video generation suffers from the problem of reference confusion. 
Distinct references with similar appearances interfere during attention, degrading identity fidelity and appearance consistency. 
To mitigate this, we augment 3D-RoPE with an additional axis that encodes \emph{side information} (e.g., which reference entities are active in a shot), thereby steering attention via structured phase modulation.

Formally, we extend the positional axes from $\boldsymbol{p}=(t,h,w)$ to $\boldsymbol{p}^{*}=(t,h,w,s)$, where the side information coordinate $s$ reflects reference entity presence. 
Suppose there are at most $K$ reference entities considered. 
For a visual token $\boldsymbol{x}$, define its side information $\boldsymbol{s}(\boldsymbol{x}) \in \{0,1\}^{K}$, where $s_i(\boldsymbol{x})=1$ indicates that reference $i$ (e.g., \texttt{@character\_i}) is present in the corresponding shot prompt, and $s_i(\boldsymbol{x})=0$ otherwise. For tokens originating from the $i$-th reference, $\boldsymbol{s}(\boldsymbol{x})$ is one-hot with $s_i(\boldsymbol{x})=1$.

We define the side information distance between tokens $\boldsymbol{x}_m$ and $\boldsymbol{x}_n$ as
\begin{equation}
\Delta^{s}_{m,n} \;=\; \big|\boldsymbol{s}(\boldsymbol{x}_m) - \boldsymbol{s}(\boldsymbol{x}_n)\big| \;\in\; \{0,1\}^{K},
\end{equation}
which quantifies per-reference agreement ($0$) or mismatch ($1$). 
Let $D_s=2K$ be the dimensionality allocated to the side-information subspace, and assume $D_t + D_h + D_w + D_s = D$. The overall SideInfo-RoPE rotation becomes

\begin{align}
    \boldsymbol{R}_{\Delta^{\boldsymbol{p}^{*}}_{m,n}} =  \{ \bigoplus_{i=1}^{\frac{D_t}{2}} \boldsymbol{R}^{(i)}_{\Delta^t_{m,n}} \} & \oplus \{ \bigoplus_{i=1}^{\frac{D_s}{2}} \hat{\boldsymbol{R}}^{(i)}_{\Delta^{s}_{m,n}} \} \notag \\ \oplus \{ \bigoplus_{i=1+\frac{D_t}{2}+\frac{D_s}{2}}^{\frac{D_t}{2}+\frac{D_s}{2}+\frac{D_h}{2}} \boldsymbol{R}^{(i)}_{\Delta^{h}_{m,n}} \} & \oplus \{ \bigoplus_{i=\frac{D_t}{2}+\frac{D_s}{2}+\frac{D_h}{2}+1}^{\frac{D}{2}} \boldsymbol{R}^{(i)}_{\Delta^{w}_{m,n}} \},
\end{align}

where $\hat{\boldsymbol{R}}^{(i)}_{\Delta^{s}_{m,n}}$ is a $2$D rotation in side information space associated with the $i^{th}$ reference image.
Concretely, define for  $i=1,\dots,K$:
\begin{align}
\hat{\boldsymbol{R}}^{(i)}_{\Delta^{s}_{m,n}} \;&=\;
\begin{bmatrix}
\cos\!\big(\phi_i \cdot {\Delta^{s}_{m,n}}^{(i)}\big) & -\sin\!\big(\phi_i \cdot {\Delta^{s}_{m,n}}^{(i)}\big) \\
\sin\!\big(\phi_i \cdot {\Delta^{s}_{m,n}}^{(i)}\big) & \phantom{-}\cos\!\big(\phi_i \cdot {\Delta^{s}_{m,n}}^{(i)}\big)
\end{bmatrix},
\\
 \phi_i \;&=\; \frac{2\pi i - \pi}{K},
\end{align}
with ${\Delta^s_{m,n}}^{(i)} \in \{0,1\}$ the $i$-th entry of $\Delta^{s}_{m,n}$.
Given that the candidate values for side information are limited, unlike the relative positions of (t, h, w) which exhibit extensive possibilities, we discretize the rotation phase offset in the side information space as uniform partitions of the $2\pi$ period.
Finally, tokens that share identical side information (${\Delta^{s}_{m,n}}^{(i)}=0$ for all $i$) experience identity-preserving, phase-aligned interactions, while tokens that disagree on entity presence (${\Delta^{s}_{m,n}}^{(i)}=1$ for some $i$) are rotated by phase offsets, attenuating cross-reference interference during attention.

\subsection{Hierarchical Cross-Attention}
\label{subsec:refmask}
To enable multi-reference, multi-shot video generation, we design a \textbf{Hierarchical Cross-Attention} that structures cross-attention in a global-local manner.
Reference tokens serve as global anchors and attend to all textual tokens, providing shared identity and style guidance across shots. 
In contrast, the video tokens of each shot attend only to the text tokens describing that shot, ensuring localized conditioning and reducing cross-shot interference.

Let the visual sequence contain \(L_v\) tokens in total, where the first \(L_{\mathrm{ref}}\) tokens correspond to the reference visual tokens and the remaining tokens correspond to generated video tokens. Let the text sequence contain \(L_t\) tokens, divided into \(S\) consecutive fixed-length segments \(\{\mathcal{T}_s\}_{s=1}^S\), each corresponding to one shot, with \(|\mathcal{T}_s|=T\). Thus, \(L_t = S \times T\). We denote by \(\mathcal{V}_s\) the set of video token indices associated with shot \(s\), such that the non-reference visual tokens are partitioned as \(\bigcup_{s=1}^S \mathcal{V}_s\).

We construct a binary mask \(\mathbf{M}\in\{0,1\}^{L_v\times L_t}\) for cross-attention as:
\[
\mathbf{M}[1{:}L_{ref},\,1{:}L_t]=1,\
\mathbf{M}[\mathcal{V}_s,\,\mathcal{T}_s]=1,\
\mathbf{M}[\mathcal{V}_s,\,\mathcal{T}_{s'\neq s}]=0.
\]

Therefore, reference tokens attend to the entire text sequence, while the video tokens of shot \(s\) attend only to their corresponding text segment \(\mathcal{T}_s\).

In practice, each \(\mathcal{T}_s\) is implemented as a fixed-length text chunk, and the video token sets \(\{\mathcal{V}_s\}_{s=1}^S\) are determined by the temporal span of each shot in the multi-shot video.
\begin{figure*}[htbp]
\begin{center}
\includegraphics[width=1.0\linewidth]{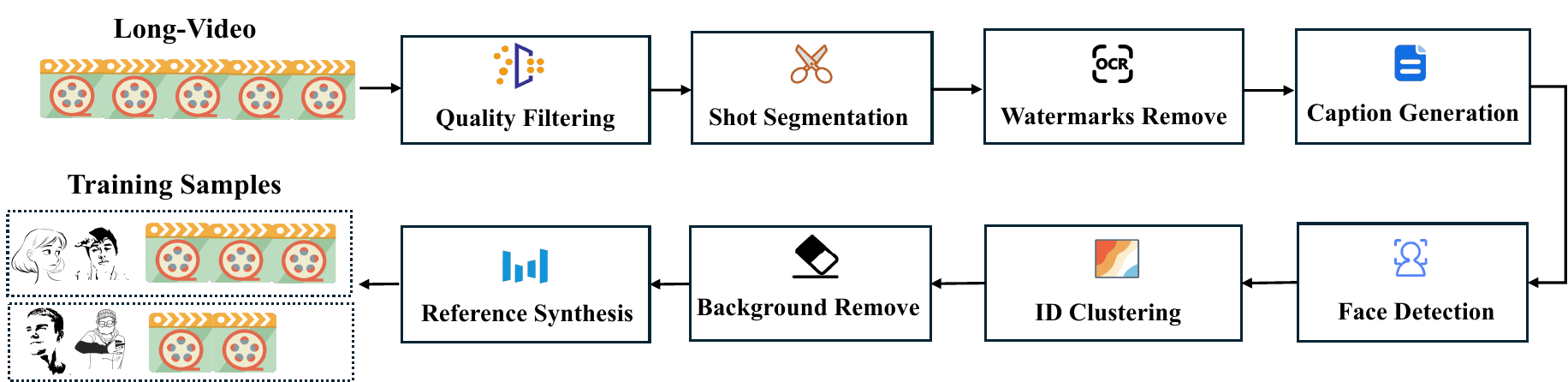}
\end{center}
\vspace{-15pt}
   \caption{
   Data pipeline for multi-reference multi-shot video generation.
The pipeline transforms raw long videos into multi-shot training samples. It includes video processing (quality filtering, shot segmentation, watermark removal, caption generation) and reference construction (face detection, ID clustering, background removal, and seedream-enhanced reference synthesis). These steps ensure clean, consistent identities and high-quality supervision for video generation.
   }
 \label{datapipeline}
\end{figure*}

\subsection{Data Pipeline}
As illustrated in Fig.~\ref{datapipeline}, we design a data pipeline that transforms raw long videos into structured multi-shot training samples with side-information annotations.
Each video is first processed by a quality filtering stage, where visual quality assessment (VQA) models evaluate aesthetics \cite{schuhmann2022laion}, sharpness, and exposure, together with heuristic checks such as border detection and frame-stability analysis. Low-quality content is discarded, and the remaining footage is segmented into single-shot clips using AutoShot \cite{zhu2023autoshot} and PySceneDetect, whose complementary sensitivities help capture both abrupt cuts and gradual transitions.
Each shot is then refined through OCR-based cropping to remove watermarks and subtitles while preserving the original aspect ratio. We further generate captions using a multimodal large language model \cite{bai1others} and merge temporally adjacent clips into multi-shot sequences, with transition frames trimmed to improve temporal continuity.

For reference construction, we perform face detection and identity clustering over all shots in each long video to group faces belonging to the same character. 
We retain only identities with sufficient temporal occurrence to provide reliable supervision for multi-shot training.
For each identity cluster, we construct two types of references: (1) a raw reference obtained directly from the clustered face crops, and (2) a seedream-enhanced reference \cite{seedream2025seedream} synthesized as a frontal and perceptually refined portrait. The clustered identity label is then propagated to the corresponding shots, yielding explicit side information (e.g., \texttt{@character1}, \texttt{@character2}) for training with SideInfo-RoPE.
The resulting dataset consists of multi-shot video segments paired with captions, dual-branch reference portraits, and per-shot identity labels, providing high-quality supervision for multi-reference multi-shot video generation.

\section{Experiment}

\begin{figure*}[htbp]
\begin{center}
\includegraphics[width=0.95\linewidth]{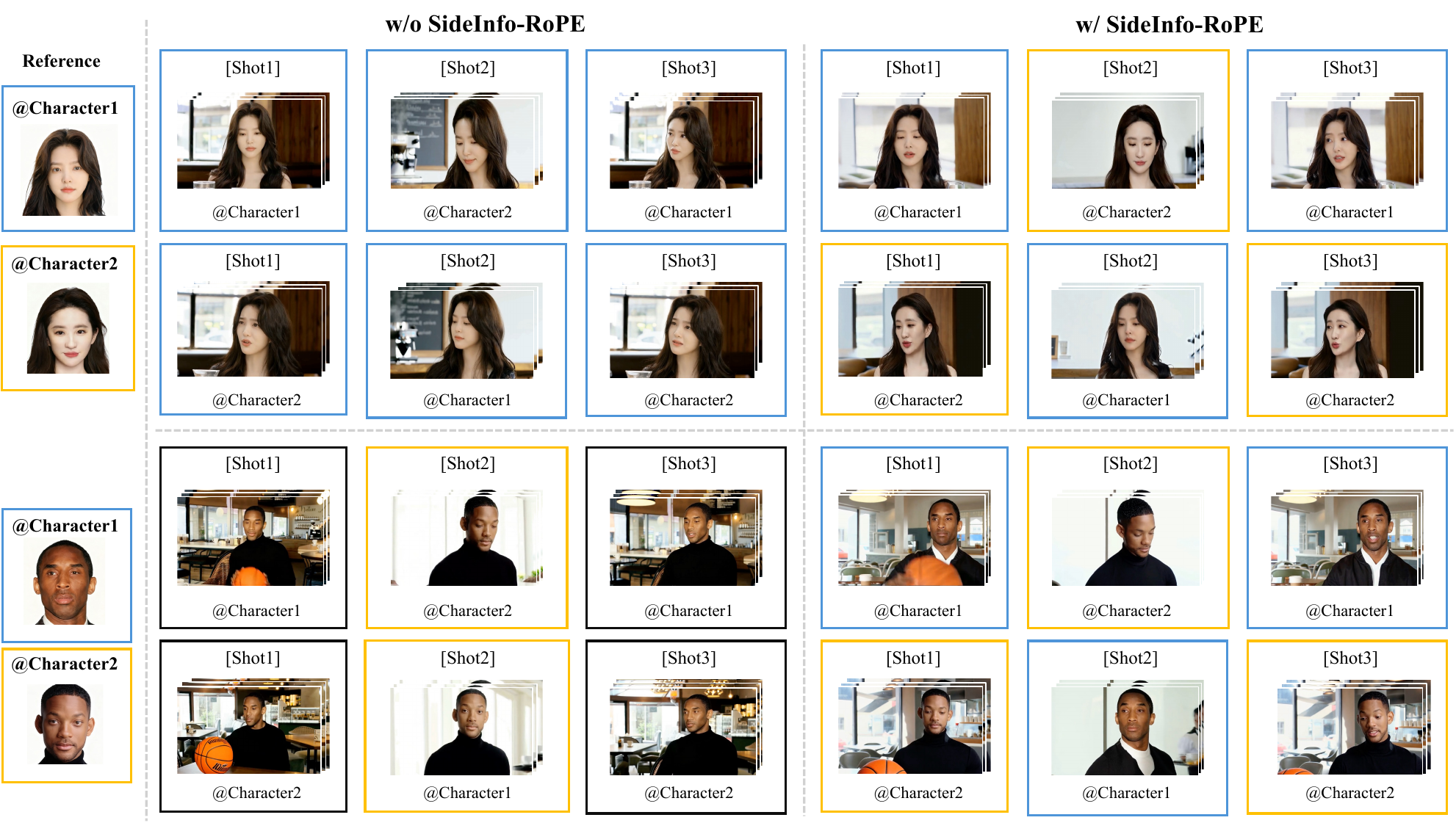}
\vspace{-18pt}
\end{center}
   \caption{
   Qualitative effect of SideInfo-RoPE on shot-level identity grounding.
We test two pairs of visually similar female and male characters using descriptions associated with two reference portraits (\texttt{@character1}, \texttt{@character2}).
Without SideInfo-RoPE, the model often exhibits incorrect or ambiguous identity grounding, including both identity swaps and failure cases that do not clearly match either reference.
With SideInfo-RoPE, the intended reference-shot correspondence is preserved more reliably.
Colored boxes indicate grounding to the corresponding reference identity, while black boxes denote ambiguous or failed grounding.
   }
 \label{fig5_ualitative_effect}
\end{figure*}

\begin{figure*}[htbp]
\begin{center}
\includegraphics[width=0.95\linewidth]{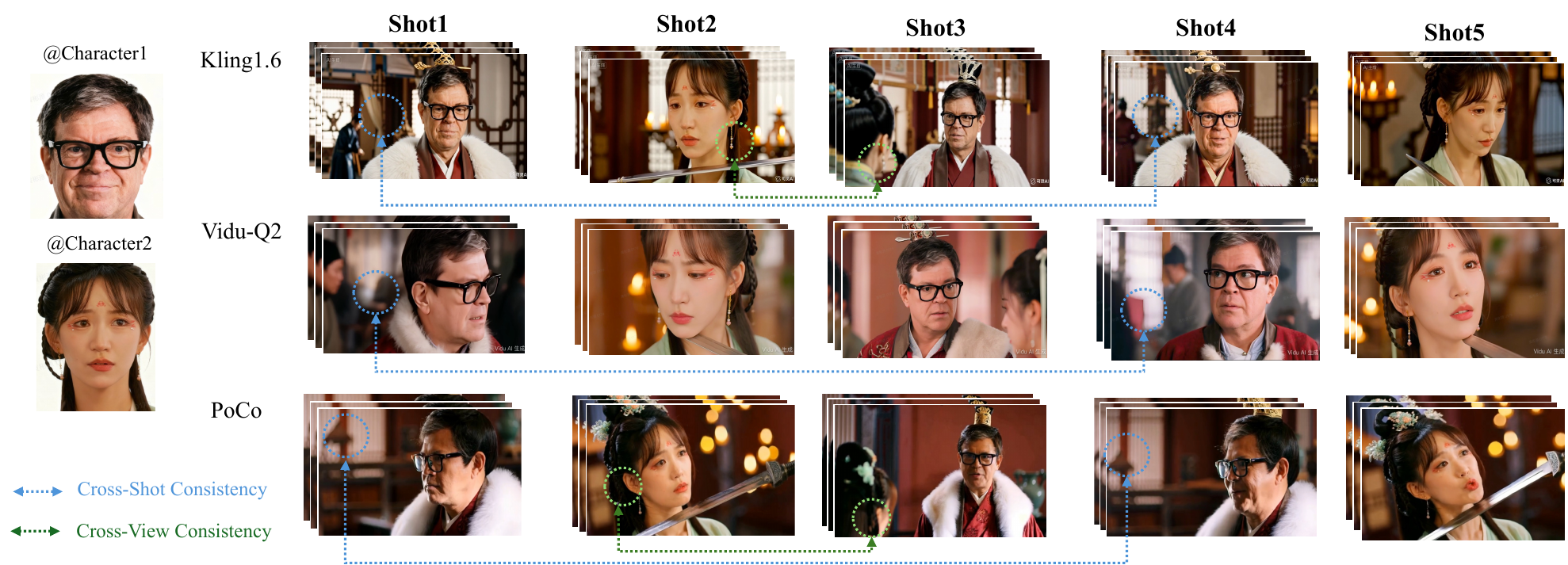}
\end{center}
\vspace{-15pt}
\caption{
Comparison with commercial reference-to-video methods under the same text prompts and identity references. Compared with Kling-1.6 and Vidu-Q2, PoCo achieves better cross-shot continuity and cross-view consistency, preserving identity, scene layout, lighting, and fine-grained appearance more faithfully.
}
 \label{comp_with_r2v}
\end{figure*}

\textbf{Experimental Setup.}
PoCo is built upon the VACE-Wan2.1-14B video generation framework~\cite{jiang2025vace}, inheriting its diffusion transformer architecture for multi-shot video generation. 
Both training and inference are conducted on 9-second videos at 480p resolution and 16 fps under a two-reference setting. 
The learning rate is set to \(1\times10^{-5}\), and four channels are allocated to encode the SideInfo axis, corresponding to two SideInfo rotation planes for the two references. 
For evaluation, we compare PoCo against several representative methods, including Phantom~\cite{liu2025phantom,chen2025phantom}, VACE~\cite{jiang2025vace}, and EchoShot~\cite{wang2025echoshot}.

\noindent\textbf{Test Set.}
Reference images are curated independently from Internet sources and model-generated content. 
Prompts are generated by GPT-5 and further refined by human annotators. 
Both images and prompts are constructed to be out-of-distribution with respect to the training data. 
We evaluate on 54 shots from 18 multi-shot videos, using 9 look-alike reference image pairs. 
All evaluations are conducted on 9-second videos.

\noindent\textbf{Evaluation Protocol and Metrics.}
We follow the metric design of OpenS2V~\cite{yuan2025opens2v} and report results in three groups: \textit{Cross-Shot}, \textit{Alignment}, and \textit{Dynamics}. 
For single-shot baselines, we generate three shots independently using the same seed and background prompt. Cross-shot metrics are then computed by aggregating results across the generated shots. 
\textbf{Cross-Shot} evaluates consistency across different shots using two metrics: 
(1) \textit{CrossShot-FaceSim}, computed by averaging CurFace cosine similarities~\cite{huang2020curricularface} between sampled frames from different shots; and 
(2) \textit{CrossShot-DINO}, computed by averaging cosine similarities between DINOv2 embeddings~\cite{oquab2024dinov2} of sampled frames from different shots to measure background semantic consistency. 
\textbf{Alignment} evaluates per-shot conditioning quality using 
(3) \textit{FaceSim}, computed by averaging CurFace cosine similarities between sampled frames and their corresponding reference images within each shot; and 
(4) \textit{Prompt}, following the OpenS2V~\cite{yuan2025opens2v} protocol, measures alignment between generated frames and textual prompts using GME~\cite{zhang2024gme}. 
\textbf{Dynamics} evaluates temporal quality using two VBench metrics~\cite{huang2024vbench}: 
(5) \textit{Smooth}, which measures motion smoothness based on motion priors from a video frame interpolation model~\cite{li2023amt}; and 
(6) \textit{Flicker}, which measures temporal flickering by computing frame differences across sampled video frames. 
For each generated multi-shot video, we uniformly sample four frames per shot and compute each metric accordingly. We report the mean score over all evaluated pairs or samples.

\subsection{Quantitative Results.}
Tab.~\ref{maintable} summarizes the quantitative comparison. 
PoCo achieves the best overall performance, obtaining the highest AvgScore under both the overall (83.46) and w/o Alignment-FaceSim (86.13) settings. 
Its main advantage lies in cross-shot consistency, where it achieves the best CrossShot-FaceSim (89.35) and CrossShot-DINO (92.66) among all methods. 
Compared with Phantom-14B, the strongest single-shot baseline, PoCo improves these metrics by \(+3.23\) and \(+19.42\), respectively; compared with the multi-shot baseline EchoShot, the gains are \(+2.30\) and \(+12.85\). 
Notably, although PoCo is built on VACE-14B, it also improves single-shot face alignment from 67.05 to 70.12. 
Overall, PoCo strengthens cross-shot coherence while maintaining strong per-shot fidelity.

\begin{table*}[t]
\centering
\caption{Quantitative comparison for multi-shot multi-reference video generation. AvgScore is computed as the arithmetic mean of the reported metric scores. “w/o Alignment-FaceSim” excludes Alignment-FaceSim, which is unavailable for EchoShot.}
\vspace{-10pt}
\setlength{\tabcolsep}{7pt}
\renewcommand{\arraystretch}{1.1}
\begin{adjustbox}{width=\textwidth,center}
\begin{tabular}{l c cc cc cc cc}
\toprule
\textbf{Method} & \textbf{Shot-Type} &
\multicolumn{2}{c}{\textbf{AvgScore}~$\uparrow$} &
\multicolumn{2}{c}{\textbf{Cross-Shot}~$\uparrow$} &
\multicolumn{2}{c}{\textbf{Alignment}~$\uparrow$} &
\multicolumn{2}{c}{\textbf{Dynamics}~$\uparrow$} \\
\cmidrule(lr){3-4}\cmidrule(lr){5-6}\cmidrule(lr){7-8}\cmidrule(lr){9-10}
& &
\multicolumn{1}{c}{Overall} &
\multicolumn{1}{c}{w/o Alignment-FaceSim} &
\multicolumn{1}{c}{FaceSim} &
\multicolumn{1}{c}{DINO} &
\multicolumn{1}{c}{FaceSim} &
\multicolumn{1}{c}{Prompt} &
\multicolumn{1}{c}{Smooth} &
\multicolumn{1}{c}{Flicker} \\
\midrule
Phantom-1.3B & Single-Shot & 80.07 & 80.00 & 80.37 & 71.15 & \textbf{80.37} & 52.72 & 98.43 & 97.35 \\
Phantom-14B  & Single-Shot & 80.72 & 82.32 & 86.12 & 73.24 & 72.75 & \textbf{55.25} & 98.74 & 98.23 \\
VACE-1.3B    & Single-Shot & 71.60 & 75.28 & 56.02 & 69.14 & 53.21 & 53.89 & 98.85 & 98.50 \\
VACE-14B     & Single-Shot & 75.56 & 77.46 & 69.49 & 67.30 & 67.05 & 52.60 & 99.00 & \textbf{98.89} \\
\midrule
EchoShot     & Multi-Shot  & N/A   & 83.82 & 87.05 & 79.81 & N/A & \textbf{54.34} & \textbf{99.09} & 98.79 \\
\midrule
\textbf{PoCo (Ours)} & Multi-Shot & \textbf{83.46} & \textbf{86.13} & \textbf{89.35} & \textbf{92.66} & 70.12 & 51.29 & 98.87 & 98.46 \\
\bottomrule
\end{tabular}
\label{maintable}
\end{adjustbox}
\end{table*}

\begin{table*}[t]
\centering
\caption{Ablation results for SideInfo-RoPE with different temporal channel configurations.}
\vspace{-10pt}
\setlength{\tabcolsep}{7pt}
\renewcommand{\arraystretch}{1.1}
\begin{adjustbox}{width=0.85\textwidth,center}
\begin{tabular}{lcccc}
\toprule
\textbf{Method}  &
\multicolumn{1}{c}{\textbf{CrossShot-FaceSim}~$\uparrow$} &
\multicolumn{1}{c}{\textbf{CrossShot-DINO}~$\uparrow$} &
\multicolumn{1}{c}{\textbf{FaceSim}~$\uparrow$} &
\multicolumn{1}{c}{\textbf{Prompt Following}~$\uparrow$} \\
\midrule
w/o SideInfo-RoPE    & 77.29 &  91.25 & 45.42 & 52.40 \\
w/ SideInfo-RoPE-Tlow  & \textbf{81.55} & \textbf{91.32}  & \textbf{60.35} & 52.96 \\
w/ SideInfo-RoPE-Thigh &  80.96 &  91.32 & 55.54  & \textbf{53.55} \\
\bottomrule
\end{tabular}
\end{adjustbox}
\label{ablationtable}
\end{table*}

\subsection{Qualitative Results.}

\textbf{Ablation on SideInfo-RoPE.}
Fig.~\ref{fig5_ualitative_effect} illustrates the effect of SideInfo-RoPE on identity grounding. We consider two pairs of visually similar characters, one female pair and one male pair. For each pair, the textual descriptions correspond to two reference portraits, denoted as \texttt{@character1} and \texttt{@character2}. Without SideInfo-RoPE, the baseline often fails to reliably ground the intended identity: the generated character may be confused with the other reference or may not faithfully match either one. In contrast, SideInfo-RoPE improves identity disambiguation between similar faces, yielding more accurate identity grounding and more consistent character assignment across shots.

\noindent\textbf{Comparison with Commercial Reference-to-Video Generators.}
Fig.~\ref{comp_with_r2v} compares PoCo with Vidu-Q2~\cite{bao2024vidu} and Kling-1.6 under the same text prompts and identity references. While both commercial systems can produce visually plausible results within individual shots, they often fail to maintain consistency across a multi-shot sequence. In particular, they struggle with cross-shot continuity in scene layout and lighting (e.g., between Shots 1 and 4) and with cross-view consistency in preserving fine-grained character attributes under viewpoint changes (e.g., between Shots 2 and 3). Such inconsistencies undermine coherent multi-shot storytelling. In contrast, PoCo preserves character identity, costume, accessories, and background more consistently across both shots and views, enabling higher-fidelity and more controllable cinematic generation.

\subsection{Ablation Study.}
We ablate SideInfo-RoPE on the VACE-14B backbone~\cite{jiang2025vace}. To enable efficient yet fair comparison, all variants are trained for half of the full schedule under the same data and optimization settings.

\begin{figure}[htbp]
\begin{center}
\includegraphics[width=0.9\linewidth]{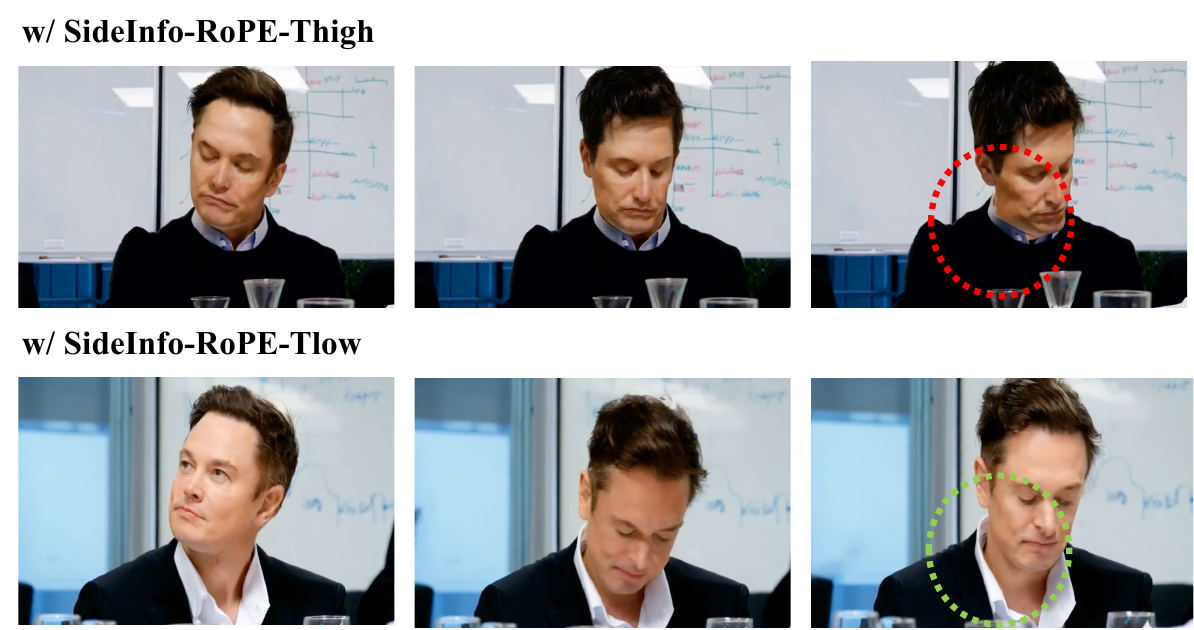}
\end{center}
\vspace{-15pt}
   \caption{
Comparison of temporal channel selection for SideInfo-RoPE. Thigh introduces facial motion artifacts and weaker identity consistency, while Tlow yields smoother motion and better identity preservation.
   }
 \label{comp-frequenc}
\end{figure}

\noindent\textbf{Effectiveness of SideInfo-RoPE.}
As shown in Tab.~\ref{ablationtable}, introducing SideInfo-RoPE consistently improves identity-related metrics. Compared with the baseline without SideInfo-RoPE, the best configuration, SideInfo-RoPE-Tlow, raises CrossShot-FaceSim from 77.29 to 81.55 and FaceSim from 45.42 to 60.35, while leaving CrossShot-DINO nearly unchanged. This suggests that SideInfo-RoPE effectively injects reference-aware guidance, helping the model associate each shot with the correct reference identity and reducing cross-shot identity confusion without harming overall visual consistency.

\noindent\textbf{Ablation on temporal channel selection.}
In Wan2.1, RoPE channels are allocated to either the temporal (T) or spatial (H/W) axes. To incorporate SideInfo-RoPE, we reassign a subset of channels to a SideInfo axis. We reallocate temporal channels rather than H/W channels, since the side information is associated with shot-wise temporal assignment. As shown in Tab.~\ref{ablationtable}, assigning low-frequency temporal channels to SideInfo-RoPE (T-low) yields the best performance, outperforming both the no-SideInfo-RoPE baseline and the high-frequency variant. In contrast, using high-frequency temporal channels (T-high) leads to weaker identity consistency and more motion artifacts, as also illustrated in Fig.~\ref{comp-frequenc}. We hypothesize that high-frequency temporal channels are more involved in modeling rapid temporal variations, so redirecting them to side information interferes more with motion modeling, whereas reallocating low-frequency channels better preserves smooth temporal dynamics while improving identity consistency.

\section{Conclusion and Limitations}
We presented PoCo, a positional-embedding-based context controller for multi-reference, multi-shot video generation. By extending RoPE with SideInfo-RoPE, PoCo introduces side information as an additional relational axis, enabling explicit reference-aware token routing while preserving the efficiency and implicit semantic consistency of the original attention mechanism. Extensive experiments show that PoCo substantially improves identity fidelity, background consistency, and cross-shot coherence over both shot-segmented R2V pipelines and existing multi-shot baselines. 

Our current design mainly addresses cross-shot reference confusion, and is less suited to fine-grained intra-shot control of multiple highly similar subjects, such as precise action binding or tightly coordinated interactions within a single frame. Extending PoCo toward denser instance-level spatial-temporal control without compromising global consistency remains an important direction for future work.

{
    \small
    \bibliographystyle{ieeenat_fullname}
    \bibliography{main}

@String(AAAI = {AAAI})

@inproceedings{liu2025phantom,
  title={Phantom: Subject-consistent video generation via cross-modal alignment},
  author={Liu, Lijie and Ma, Tianxiang and Li, Bingchuan and Chen, Zhuowei and Liu, Jiawei and Li, Gen and Zhou, Siyu and He, Qian and Wu, Xinglong},
  booktitle={Proceedings of the IEEE/CVF International Conference on Computer Vision},
  year={2025}
}

@inproceedings{jiang2025vace,
  title={Vace: All-in-one video creation and editing},
  author={Jiang, Zeyinzi and Han, Zhen and Mao, Chaojie and Zhang, Jingfeng and Pan, Yulin and Liu, Yu},
  booktitle={Proceedings of the IEEE/CVF International Conference on Computer Vision},
  year={2025}
}

@article{liu2023cones,
  title={Cones: Concept neurons in diffusion models for customized generation},
  author={Liu, Zhiheng and Feng, Ruili and Zhu, Kai and Zhang, Yifei and Zheng, Kecheng and Liu, Yu and Zhao, Deli and Zhou, Jingren and Cao, Yang},
  journal={arXiv preprint arXiv:2303.05125},
  year={2023}
}

@article{wang2024instantid,
  title={Instantid: Zero-shot identity-preserving generation in seconds},
  author={Wang, Qixun and Bai, Xu and Wang, Haofan and Qin, Zekui and Chen, Anthony and Li, Huaxia and Tang, Xu and Hu, Yao},
  journal={arXiv preprint arXiv:2401.07519},
  year={2024}
}

@article{guo2024pulid,
  title={Pulid: Pure and lightning id customization via contrastive alignment},
  author={Guo, Zinan and Wu, Yanze and Chen, Zhuowei and Chen, Lang and Zhang, Peng and He, Qian},
  journal={Advances in neural information processing systems},
  year={2024}
}

@article{zhang2024gme,
  title={GME: improving universal multimodal retrieval by multimodal LLMs},
  author={Zhang, Xin and Zhang, Yanzhao and Xie, Wen and Li, Mingxin and Dai, Ziqi and Long, Dingkun and Xie, Pengjun and Zhang, Meishan and Li, Wenjie and Zhang, Min},
  journal={arXiv preprint arXiv:2412.16855},
  year={2024}
}

@inproceedings{huang2024vbench,
  title={Vbench: Comprehensive benchmark suite for video generative models},
  author={Huang, Ziqi and He, Yinan and Yu, Jiashuo and Zhang, Fan and Si, Chenyang and Jiang, Yuming and Zhang, Yuanhan and Wu, Tianxing and Jin, Qingyang and Chanpaisit, Nattapol and others},
  booktitle={Proceedings of the IEEE/CVF Conference on Computer Vision and Pattern Recognition},
  year={2024}
}

@article{yuan2025opens2v,
  title={Opens2v-nexus: A detailed benchmark and million-scale dataset for subject-to-video generation},
  author={Yuan, Shenghai and He, Xianyi and Deng, Yufan and Ye, Yang and Huang, Jinfa and Lin, Bin and Luo, Jiebo and Yuan, Li},
  journal={arXiv preprint arXiv:2505.20292},
  year={2025}
}

@article{wang2023videocomposer,
  title={Videocomposer: Compositional video synthesis with motion controllability},
  author={Wang, Xiang and Yuan, Hangjie and Zhang, Shiwei and Chen, Dayou and Wang, Jiuniu and Zhang, Yingya and Shen, Yujun and Zhao, Deli and Zhou, Jingren},
  journal={Advances in Neural Information Processing Systems},
  year={2023}
}

@article{huang2023composer,
  title={Composer: Creative and controllable image synthesis with composable conditions},
  author={Huang, Lianghua and Chen, Di and Liu, Yu and Shen, Yujun and Zhao, Deli and Zhou, Jingren},
  journal={arXiv preprint arXiv:2302.09778},
  year={2023}
}

@article{zhao2025controlvideo,
  title={ControlVideo: conditional control for one-shot text-driven video editing and beyond},
  author={Zhao, Min and Wang, Rongzhen and Bao, Fan and Li, Chongxuan and Zhu, Jun},
  journal={Science China Information Sciences},
  year={2025}
}

@inproceedings{huang2020curricularface,
  title={Curricularface: adaptive curriculum learning loss for deep face recognition},
  author={Huang, Yuge and Wang, Yuhan and Tai, Ying and Liu, Xiaoming and Shen, Pengcheng and Li, Shaoxin and Li, Jilin and Huang, Feiyue},
  booktitle={proceedings of the IEEE/CVF conference on computer vision and pattern recognition},
  year={2020}
}

@article{bai1others,
  title={Qwen2.5-vl technical report},
  author={Bai, Shuai and Chen, Keqin and Liu, Xuejing and Wang, Jialin and Ge, Wenbin and Song, Sibo and Dang, Kai and Wang, Peng and Wang, Shijie and Tang, Jun and others},
  journal={arXiv preprint arXiv:2502.13923},
  year={2025}
}

@article{xue2025stand,
  title={Stand-in: A lightweight and plug-and-play identity control for video generation},
  author={Xue, Bowen and Duan, Zheng-Peng and Yan, Qixin and Wang, Wenjing and Liu, Hao and Guo, Chun-Le and Li, Chongyi and Li, Chen and Lyu, Jing},
  journal={arXiv preprint arXiv:2508.07901},
  year={2025}
}

@article{schuhmann2022laion,
  title={Laion-5b: An open large-scale dataset for training next generation image-text models},
  author={Schuhmann, Christoph and Beaumont, Romain and Vencu, Richard and Gordon, Cade and Wightman, Ross and Cherti, Mehdi and Coombes, Theo and Katta, Aarush and Mullis, Clayton and Wortsman, Mitchell and others},
  journal={Advances in neural information processing systems},
  year={2022}
}

@inproceedings{zhu2023autoshot,
  title={Autoshot: A short video dataset and state-of-the-art shot boundary detection},
  author={Zhu, Wentao and Huang, Yufang and Xie, Xiufeng and Liu, Wenxian and Deng, Jincan and Zhang, Debing and Wang, Zhangyang and Liu, Ji},
  booktitle={Proceedings of the IEEE/CVF Conference on Computer Vision and Pattern Recognition},
  year={2023}
}

@inproceedings{li2023amt,
  title={Amt: All-pairs multi-field transforms for efficient frame interpolation},
  author={Li, Zhen and Zhu, Zuo-Liang and Han, Ling-Hao and Hou, Qibin and Guo, Chun-Le and Cheng, Ming-Ming},
  booktitle={Proceedings of the IEEE/CVF Conference on Computer Vision and Pattern Recognition},
  year={2023}
}

@article{seedream2025seedream,
  title={Seedream 4.0: Toward next-generation multimodal image generation},
  author={Seedream, Team and Chen, Yunpeng and Gao, Yu and Gong, Lixue and Guo, Meng and Guo, Qiushan and Guo, Zhiyao and Hou, Xiaoxia and Huang, Weilin and Huang, Yixuan and others},
  journal={arXiv preprint arXiv:2509.20427},
  year={2025}
}

@inproceedings{huang2025subjectdrive,
  title={Subjectdrive: Scaling generative data in autonomous driving via subject control},
  author={Huang, Binyuan and Wen, Yuqing and Zhao, Yucheng and Hu, Yaosi and Liu, Yingfei and Jia, Fan and Mao, Weixin and Wang, Tiancai and Zhang, Chi and Chen, Chang Wen and others},
  booktitle={Proceedings of the AAAI Conference on Artificial Intelligence},
  year={2025}
}

@article{chen2025phantom,
  title={Phantom-data: Towards a general subject-consistent video generation dataset},
  author={Chen, Zhuowei and Li, Bingchuan and Ma, Tianxiang and Liu, Lijie and Liu, Mingcong and Zhang, Yi and Li, Gen and Li, Xinghui and Zhou, Siyu and He, Qian and others},
  journal={arXiv preprint arXiv:2506.18851},
  year={2025}
}

@article{bao2024vidu,
  title={Vidu: a highly consistent, dynamic and skilled text-to-video generator with diffusion models},
  author={Bao, Fan and Xiang, Chendong and Yue, Gang and He, Guande and Zhu, Hongzhou and Zheng, Kaiwen and Zhao, Min and Liu, Shilong and Wang, Yaole and Zhu, Jun},
  journal={arXiv preprint arXiv:2405.04233},
  year={2024}
}

@article{oquab2024dinov2,
  title={DINOv2: Learning Robust Visual Features without Supervision},
  author={Oquab, Maxime and Darcet, Timoth{\'e}e and Moutakanni, Th{\'e}o and Vo, Huy and Szafraniec, Marc and Khalidov, Vasil and Fernandez, Pierre and Haziza, Daniel and Massa, Francisco and El-Nouby, Alaaeldin and others},
  journal={Transactions on Machine Learning Research Journal},
  year={2024}
}

@inproceedings{zhang2025magicmirror,
  title={Magicmirror: Id-preserved video generation in video diffusion transformers},
  author={Zhang, Yuechen and Liu, Yaoyang and Xia, Bin and Peng, Bohao and Yan, Zexin and Lo, Eric and Jia, Jiaya},
  booktitle={Proceedings of the IEEE/CVF International Conference on Computer Vision},
  year={2025}
}

@article{wu2025vmoba,
  title={VMoBA: Mixture-of-Block Attention for Video Diffusion Models},
  author={Wu, Jianzong and Hou, Liang and Yang, Haotian and Tao, Xin and Tian, Ye and Wan, Pengfei and Zhang, Di and Tong, Yunhai},
  journal={arXiv preprint arXiv:2506.23858},
  year={2025}
}

@inproceedings{liu2024video,
  title={Video-p2p: Video editing with cross-attention control},
  author={Liu, Shaoteng and Zhang, Yuechen and Li, Wenbo and Lin, Zhe and Jia, Jiaya},
  booktitle={Proceedings of the IEEE/CVF Conference on Computer Vision and Pattern Recognition},
  year={2024}
}

@article{liew2023magicedit,
  title={Magicedit: High-fidelity and temporally coherent video editing},
  author={Liew, Jun Hao and Yan, Hanshu and Zhang, Jianfeng and Xu, Zhongcong and Feng, Jiashi},
  journal={arXiv preprint arXiv:2308.14749},
  year={2023}
}

@article{qiu2025animeshooter,
  title={AnimeShooter: A Multi-Shot Animation Dataset for Reference-Guided Video Generation},
  author={Qiu, Lu and Li, Yizhuo and Ge, Yuying and Ge, Yixiao and Shan, Ying and Liu, Xihui},
  journal={arXiv preprint arXiv:2506.03126},
  year={2025}
}

@article{wang2025echoshot,
  title={EchoShot: Multi-Shot Portrait Video Generation},
  author={Wang, Jiahao and Sheng, Hualian and Cai, Sijia and Zhang, Weizhan and Yan, Caixia and Feng, Yachuang and Deng, Bing and Ye, Jieping},
  journal={arXiv preprint arXiv:2506.15838},
  year={2025}
}

@article{wan2025wan,
  title={Wan: Open and advanced large-scale video generative models},
  author={Wan, Team and Wang, Ang and Ai, Baole and Wen, Bin and Mao, Chaojie and Xie, Chen-Wei and Chen, Di and Yu, Feiwu and Zhao, Haiming and Yang, Jianxiao and others},
  journal={arXiv preprint arXiv:2503.20314},
  year={2025}
}

@article{yang2024cogvideox,
  title={Cogvideox: Text-to-video diffusion models with an expert transformer},
  author={Yang, Zhuoyi and Teng, Jiayan and Zheng, Wendi and Ding, Ming and Huang, Shiyu and Xu, Jiazheng and Yang, Yuanming and Hong, Wenyi and Zhang, Xiaohan and Feng, Guanyu and others},
  journal={arXiv:2408.06072},
  year={2024}
}

@article{cai2025moc,
  title={Mixture of contexts for long video generation},
  author={Cai, Shengqu and Yang, Ceyuan and Zhang, Lvmin and Guo, Yuwei and Xiao, Junfei and Yang, Ziyan and Xu, Yinghao and Yang, Zhenheng and Yuille, Alan and Guibas, Leonidas and others},
  journal={arXiv preprint arXiv:2508.21058},
  year={2025}
}

@inproceedings{yu2025context,
  title={Context as memory: Scene-consistent interactive long video generation with memory retrieval},
  author={Yu, Jiwen and Bai, Jianhong and Qin, Yiran and Liu, Quande and Wang, Xintao and Wan, Pengfei and Zhang, Di and Liu, Xihui},
  booktitle={Proceedings of the SIGGRAPH Asia 2025 Conference Papers},
  year={2025}
}

@article{jia2025moga,
  title={MoGA: Mixture-of-Groups Attention for End-to-End Long Video Generation},
  author={Jia, Weinan and Lu, Yuning and Huang, Mengqi and Wang, Hualiang and Huang, Binyuan and Chen, Nan and Liu, Mu and Jiang, Jidong and Mao, Zhendong},
  journal={arXiv preprint arXiv:2510.18692},
  year={2025}
}

@misc{openai@sora2,
  author = {{OpenAI}},
  title = {Sora 2},
  year = {2025},
  howpublished = {\url{https://openai.com/index/sora-2/}},
  note = {Accessed: 2026-03-27}
}

@article{hu2025hunyuancustom,
  title={Hunyuancustom: A multimodal-driven architecture for customized video generation},
  author={Hu, Teng and Yu, Zhentao and Zhou, Zhengguang and Liang, Sen and Zhou, Yuan and Lin, Qin and Lu, Qinglin},
  journal={arXiv preprint arXiv:2505.04512},
  year={2025}
}

@article{fei2025skyreels,
  title={Skyreels-a2: Compose anything in video diffusion transformers},
  author={Fei, Zhengcong and Li, Debang and Qiu, Di and Wang, Jiahua and Dou, Yikun and Wang, Rui and Xu, Jingtao and Fan, Mingyuan and Chen, Guibin and Li, Yang and others},
  journal={arXiv preprint arXiv:2504.02436},
  year={2025}
}

@article{su2024roformer,
  title={Roformer: Enhanced transformer with rotary position embedding},
  author={Su, Jianlin and Ahmed, Murtadha and Lu, Yu and Pan, Shengfeng and Bo, Wen and Liu, Yunfeng},
  journal={Neurocomputing},
  year={2024},
}

@article{kong2024hunyuanvideo,
  title={Hunyuanvideo: A systematic framework for large video generative models},
  author={Kong, Weijie and Tian, Qi and Zhang, Zijian and Min, Rox and Dai, Zuozhuo and Zhou, Jin and Xiong, Jiangfeng and Li, Xin and Wu, Bo and Zhang, Jianwei and others},
  journal={arXiv preprint arXiv:2412.03603},
  year={2024}
}

@article{gao2025seedance,
  title={Seedance 1.0: Exploring the Boundaries of Video Generation Models},
  author={Gao, Yu and Guo, Haoyuan and Hoang, Tuyen and Huang, Weilin and Jiang, Lu and Kong, Fangyuan and Li, Huixia and Li, Jiashi and Li, Liang and Li, Xiaojie and others},
  journal={arXiv preprint arXiv:2506.09113},
  year={2025}
}

@inproceedings{kara2025shotadapter,
  title={Shotadapter: Text-to-multi-shot video generation with diffusion models},
  author={Kara, Ozgur and Singh, Krishna Kumar and Liu, Feng and Ceylan, Duygu and Rehg, James M and Hinz, Tobias},
  booktitle={Proceedings of the Computer Vision and Pattern Recognition Conference},
  year={2025}
}

@article{guo2025long,
  title={Long context tuning for video generation},
  author={Guo, Yuwei and Yang, Ceyuan and Yang, Ziyan and Ma, Zhibei and Lin, Zhijie and Yang, Zhenheng and Lin, Dahua and Jiang, Lu},
  journal={arXiv preprint arXiv:2503.10589},
  year={2025}
}

@inproceedings{chen2025multi,
  title={Multi-subject open-set personalization in video generation},
  author={Chen, Tsai-Shien and Siarohin, Aliaksandr and Menapace, Willi and Fang, Yuwei and Lee, Kwot Sin and Skorokhodov, Ivan and Aberman, Kfir and Zhu, Jun-Yan and Yang, Ming-Hsuan and Tulyakov, Sergey},
  booktitle={Proceedings of the Computer Vision and Pattern Recognition Conference},
  year={2025}
}

@article{li2023videogen,
  title={Videogen: A reference-guided latent diffusion approach for high definition text-to-video generation},
  author={Li, Xin and Chu, Wenqing and Wu, Ye and Yuan, Weihang and Liu, Fanglong and Zhang, Qi and Li, Fu and Feng, Haocheng and Ding, Errui and Wang, Jingdong},
  journal={arXiv preprint arXiv:2309.00398},
  year={2023}
}

@article{zhang2025waver,
  title={Waver: Wave your way to lifelike video generation},
  author={Zhang, Yifu and Yang, Hao and Zhang, Yuqi and Hu, Yifei and Zhu, Fengda and Lin, Chuang and Mei, Xiaofeng and Jiang, Yi and Peng, Bingyue and Yuan, Zehuan},
  journal={arXiv preprint arXiv:2508.15761},
  year={2025}
}
}


\end{document}